\newif\ifJOURNAL
\JOURNALfalse
\newif\ifWP
\WPfalse
\newif\ifarXiv
\arXivfalse

\arXivtrue

\newif\ifnotJOURNAL	
\notJOURNALtrue
\ifJOURNAL\notJOURNALfalse\fi

\newif\ifLATIN		

\ifJOURNAL
  \documentclass{cja4}

  \copyrightyear{2006}
  \vol{00}
  \issue{0}
  \DOI{000}
  \usepackage{amsmath,amsfonts,latexsym,graphicx}
  \LATINfalse
\fi

\ifWP
  \documentclass[toc]{kpnsarticle}
  \usepackage{amsmath,amsfonts,latexsym,graphicx,epsfig}
  \LATINfalse
\fi

\ifarXiv
  \documentclass{article}
  \usepackage{amsmath,amsfonts,latexsym,graphicx}
  \LATINtrue
\fi

\newif\ifnotLATIN	
\notLATINtrue
\ifLATIN\notLATINfalse\fi

\allowdisplaybreaks[3]

\ifnotLATIN
  \usepackage{CJK}
  \input{OT2enc.def}
  \newenvironment{cyr}
  {\fontencoding{OT2}\fontfamily{wncyr}\fontseries{m}\fontshape{n}\selectfont}
  {\fontencoding{OT1}\fontfamily{tir}\selectfont}
\fi

\newcommand{\bbbr}{{\mathbb{R}}}
\newcommand{\bbbn}{{\mathbb{N}}}
\newcommand{\st}{:}
\newcommand{\given}{\mathbin{|}}

\newlength{\picturewidth}
\ifJOURNAL
\fi
\ifnotJOURNAL
\fi

\newcommand{\bbbe}{{\mathbb{E}}}		
\newcommand{\Expect}{\mathop{\bbbe}\nolimits}

\newcommand{\Err}{\mathop{{\rm Err}}\nolimits}
\newcommand{\err}{\mathop{{\rm err}}\nolimits}

\newcommand{\Mult}{\mathop{{\rm Mult}}\nolimits}
\newcommand{\mult}{\mathop{{\rm mult}}\nolimits}

\newcommand{\Emp}{\mathop{{\rm Emp}}\nolimits}
\newcommand{\emp}{\mathop{{\rm emp}}\nolimits}

\ifnotJOURNAL

  \newtheorem{theorem}{Theorem}
  
\fi

\newenvironment{remark*}
  {\trivlist\item[\hskip\labelsep{\bfseries Remark}]\relax}
  {\endtrivlist}
\newenvironment{definition*}
  {\trivlist\item[\hskip\labelsep{\bfseries Definition}]\relax}
  {\endtrivlist}

\ifWP
  \title{Hedging Predictions in Machine Learning}
  \author{Alexander Gammerman and Vladimir Vovk}
  
\fi

\ifarXiv
  \title{Hedging Predictions in Machine Learning}
  \author{Alexander Gammerman and Vladimir Vovk\\
      Computer Learning Research Centre\\
      Department of Computer Science\\
      Royal Holloway, University of London\\
      Egham, Surrey TW20 0EX, UK\\
      \texttt{\{alex,vovk\}@cs.rhul.ac.uk}}
\fi

\begin{document}
\ifJOURNAL
  \title[Hedging Predictions]{Hedging Predictions\\in Machine Learning}
  \author{Alexander Gammerman}
  \author{Vladimir Vovk}
  \affiliation{Computer Learning Research Centre,
    Royal Holloway, University of London\\
    Egham, Surrey TW20 0EX}
  \email{\{alex,vovk\}@cs.rhul.ac.uk}

  \shortauthors{A.~Gammerman and V.~Vovk}

  \received{00 Month 2006}
  \revised{00 Month 2006}
\fi

\ifnotJOURNAL
  \maketitle
\fi

\begin{abstract}
  Recent advances in machine learning make it possible
  to design efficient prediction algorithms for data sets with huge numbers of parameters.
  This paper describes a new technique for ``hedging'' the predictions
  output by many such algorithms,
  including support vector machines, kernel ridge regression, kernel nearest neighbours,
  and by many other state-of-the-art methods.
  The hedged predictions for the labels of new objects
  include quantitative measures of their own accuracy and reliability.
  These measures are provably valid under the assumption of randomness,
  traditional in machine learning:
  the objects and their labels are assumed to be generated independently
  from the same probability distribution.
  In particular, it becomes possible to control (up to statistical fluctuations)
  the number of erroneous predictions by selecting a suitable confidence level.
  Validity being achieved automatically,
  the remaining goal of hedged prediction is efficiency:
  taking full account of the new objects' features
  and other available information to produce as accurate predictions as possible.
  This can be done successfully using the powerful machinery of modern machine learning.
\end{abstract}

\ifJOURNAL

  \maketitle
\fi

\section{Introduction}
\label{sec:introduction}


The two main varieties of the problem of prediction,
classification and regression,
are standard subjects in statistics and machine learning.
The classical classification and regression techniques
can deal successfully with conventional small-scale, low-dimensional data sets;
however, attempts to apply these techniques to modern high-dimensional and high-throughput data sets
encounter serious conceptual and computational difficulties.
Several new techniques,
first of all support vector machines \cite{vapnik:1995,vapnik:1998}
and other kernel methods,
have been developed in machine learning recently
with the explicit goal of dealing with high-dimensional data sets
with large numbers of objects.

A typical drawback of the new techniques is the lack of useful measures of confidence
in their predictions.
For example, some of the tightest upper bounds of the popular PAC theory
on the probability of error exceed~1 even for relatively clean data sets
(\cite{vovk/etal:2005}, p.~249).
This paper describes an efficient way to ``hedge'' the predictions
produced by the new and traditional machine-learning methods,
i.e., to complement them with measures of their accuracy and reliability.
Appropriately chosen,
not only are these measures valid and informative,
but they also take full account of the special features
of the object to be predicted.

We call our algorithms for producing hedged predictions ``conformal predictors'';
they are formally introduced in Section \ref{sec:conformal}.
Their most important property is the automatic validity under the randomness assumption
(to be discussed shortly).
Informally, validity means that conformal predictors never overrate
the accuracy and reliability of their predictions.
This property, stated in Sections \ref{sec:conformal} and \ref{sec:on-line},
is formalized in terms of finite data sequences,
without any recourse to asymptotics.

The claim of validity of conformal predictors
depends on an assumption that is shared by many other algorithms in machine learning,
which we call the assumption of randomness:
the objects and their labels are assumed to be generated independently
from the same probability distribution.
Admittedly, this is a strong assumption,
and areas of machine learning are emerging
that rely on other assumptions
(such as the Markovian assumption of reinforcement learning;
see, e.g., \cite{sutton/barto:1998})
or dispense with any stochastic assumptions altogether
(competitive on-line learning;
see, e.g., \cite{cesabianchi/lugosi:2006,vovk:2001}).
It is, however, much weaker than assuming a parametric statistical model,
sometimes complemented with a prior distribution on the parameter space,
which is customary in the statistical theory of prediction.
And taking into account the strength of the guarantees that can be proved
under this assumption,
it does not appear overly restrictive.

So we know that conformal predictors tell the truth.
Clearly, this is not enough:
truth can be uninformative and so useless.
We will refer to various measures of informativeness of conformal predictors
as their ``efficiency''.
As conformal predictors are provably valid,
efficiency is the only thing we need to worry about
when designing conformal predictors
for solving specific problems.
Virtually any classification or regression algorithm
can be transformed into a conformal predictor,
and so most of the arsenal of methods of modern machine learning
can be brought to bear on the design of efficient conformal predictors.

We start the main part of the paper, in Section \ref{sec:ideal},
with the description of an idealized predictor
based on Kolmogorov's algorithmic theory of randomness.
This ``universal predictor'' produces the best possible hedged predictions
but, unfortunately, is noncomputable.
We can, however, set ourselves the task of approximating the universal predictor
as well as possible.

In Section \ref{sec:conformal} we formally introduce the notion of conformal predictors
and state a simple result about their validity.
In that section we also briefly describe results of computer experiments
demonstrating the methodology of conformal prediction.

In Section \ref{sec:Bayesian} we consider an example demonstrating
how conformal predictors react to the violation of our model
of the stochastic mechanism generating the data
(within the framework of the randomness assumption).
If the model coincides with the actual stochastic mechanism,
we can construct an optimal conformal predictor,
which turns out to be almost as good as the Bayes-optimal confidence predictor
(the formal definitions will be given later).
When the stochastic mechanism significantly deviates from the model,
conformal predictions remain valid but their efficiency inevitably suffers.
The Bayes-optimal predictor starts producing very misleading results
which superficially look as good as when the model is correct.

In Section \ref{sec:on-line} we describe the ``on-line'' setting
of the problem of prediction,
and in Section \ref{sec:slow} contrast it with the more standard ``batch'' setting.
The notion of validity introduced in Section \ref{sec:conformal}
is applicable to both settings,
but in the on-line setting it can be strengthened:
we can now prove that the percentage of the erroneous predictions
will be close, with high probability,
to a chosen confidence level.
For the batch setting,
the stronger property of validity for conformal predictors
remains an empirical fact.
In Section \ref{sec:slow} we also discuss limitations of the on-line setting
and introduce new settings intermediate between on-line and batch.
To a large degree,
conformal predictors still enjoy the stronger property of validity
for the intermediate settings.

Section \ref{sec:induction-transduction} is devoted
to the discussion of the difference between two kinds of inference from empirical data,
induction and transduction
(emphasized by Vladimir Vapnik \cite{vapnik:1995,vapnik:1998}).
Conformal predictors belong to transduction,
but combining them with elements of induction
can lead to a significant improvement in their computational efficiency
(Section \ref{sec:ICP}).

We show how some popular methods of machine learning
can be used as underlying algorithms for hedged prediction.
We do not give the full description of these methods
and refer the reader to the existing readily accessible descriptions.
This paper is, however, self-contained in the sense
that we explain all features of the underlying algorithms
that are used in hedging their predictions.
We hope that the information we provide will enable the reader
to apply our hedging techniques
to their favourite machine-learning methods.

\section{Ideal hedged predictions}
\label{sec:ideal}


The most basic problem of machine learning is perhaps the following.
We are given a \emph{training set} of \emph{examples}
\begin{equation}\label{eq:training-set}
  (x_1,y_1),\ldots,(x_l,y_l),
\end{equation}
each example $(x_i,y_i)$, $i=1,\ldots,l$, consisting of an \emph{object} $x_i$
(typically, a vector of attributes)
and its label $y_i$;
the problem is to predict the label $y_{l+1}$
of a new object $x_{l+1}$.
Two important special cases are where the labels are known \emph{a priori}
to belong to a relatively small finite set
(the problem of \emph{classification})
and where the labels are allowed to be any real numbers
(the problem of \emph{regression}).

The usual goal of classification is to produce a prediction $\hat y_{l+1}$
that is likely to coincide with the true label $y_{l+1}$,
and the usual goal of regression is to produce a prediction $\hat y_{l+1}$
that is likely to be close to the true label $y_{l+1}$.
In the case of classification,
our goal will be to complement the prediction $\hat y_{l+1}$
with some measure of its reliability.
In the case of regression,
we would like to have some measure of accuracy and reliability of our prediction.
There is a clear trade-off between accuracy and reliability:
we can improve the former by relaxing the latter
and vice versa.
We are looking for algorithms that achieve the best possible trade-off
and for a measure that would quantify the achieved trade-off.

Let us start from the case of classification.
The idea is to try every possible label $Y$ as a candidate for $x_{l+1}$'s label
and see how well the resulting sequence
\begin{equation}\label{eq:completion}
  (x_1,y_1),\dots,(x_l,y_l),(x_{l+1},Y)
\end{equation}
conforms to the randomness assumption
(if it does conform to this assumption, we will say that it is ``random'';
this will be formalized later in this section).
The ideal case is where all $Y$s but one lead to sequences (\ref{eq:completion})
that are not random;
we can then use the remaining $Y$ as a confident prediction for $y_{l+1}$.

In the case of regression,
we can output the set of all $Y$s that lead to random (\ref{eq:completion})
as our ``prediction set''.
An obvious obstacle is that the set of all possible $Y$s is infinite
and so we cannot go through all the $Y$s explicitly,
but we will see in the next section that there are ways to overcome this difficulty.

We can see that the problem of hedged prediction
is intimately connected with the problem of testing randomness.
Different versions of the ``universal'' notion of randomness
were defined by Kolmogorov, Martin-L\"of and Levin (see, e.g., \cite{li/vitanyi:1997})
based on the existence of universal Turing machines.
Adapted to our current setting,
Martin-L\"of's definition is as follows.
Let $\mathbf{Z}$ be the set of all possible examples;
as each example consists of an object and a label,
$\mathbf{Z}=\mathbf{X}\times\mathbf{Y}$,
where $\mathbf{X}$ is the set of all possible objects
and $\mathbf{Y}$, $\left|\mathbf{Y}\right|>1$, is the set of all possible labels.
We will use $\mathbf{Z}^*$ as the notation for all finite sequences of examples.
A function $t:\mathbf{Z}^*\to[0,1]$
is a \emph{randomness test} if
\begin{enumerate}
\item
  for all $\epsilon\in(0,1)$, all $n\in\{1,2,\dots\}$
  and all probability distributions $P$ on $\mathbf{Z}$,
  \begin{equation}\label{eq:test-validity}
    P^n
    \left\{
      z\in\mathbf{Z}^n
      \st
      t(z)\le\epsilon
    \right\}
    \le
    \epsilon;
  \end{equation}
\item
  $t$ is upper semicomputable.
\end{enumerate}
The first condition means that the randomness test is required to be valid:
if, for example, we observe $t(z)\le1\%$ for our data set $z$,
then either the data set was not generated independently from the same probability distribution $P$
or a rare (of probability at most 1\%, under any $P$) event has occurred.
The second condition means that
we should be able to compute the test, in a weak sense
(we cannot require computability in the usual sense,
since the universal test can only be upper semicomputable:
it can work forever to discover \emph{all} patterns in the data sequence
that make it non-random).
Martin-L\"of (developing Kolmogorov's earlier ideas) proved
that there exists a smallest, to within a constant factor,
randomness test.

Let us fix a smallest randomness test,
call it the \emph{universal test},
and call the value it takes on a data sequence
the \emph{randomness level} of this sequence.
A random sequence is one whose randomness level is not small;
this is rather informal,
but it is clear that for finite data sequences we cannot have a clear-cut division
of all sequences into random and non-random
(like the one defined by Martin-L\"of \cite{martin-lof:1966} for infinite sequences).
If $t$ is a randomness test, not necessarily universal,
the value that it takes on a data sequence will be called
the \emph{randomness level detected by} $t$.

\begin{remark*}
  The word ``random'' is used in (at least) two different senses in the existing literature.
  In this paper we need both but, luckily,
  the difference does not matter within our current framework.
  First, randomness can refer to the assumption that the examples
  are generated independently from the same distribution;
  this is the origin of our ``assumption of randomness''.
  Second, a data sequence is said to be random with respect to a statistical model
  if the universal test (a generalization of the notion of universal test as defined above)
  does not detect any lack of conformity between the two.
  Since the only statistical model we are interested in this paper
  is the one embodying the assumption of randomness,
  we have a perfect agreement between the two senses.
\end{remark*}

\subsection*{Prediction with Confidence and Credibility}

Once we have a randomness test $t$, universal or not,
we can use it for hedged prediction.
There are two natural ways to package the results
of such predictions:
in this subsection we will describe the way that can only be used
in classification problems.
If the randomness test is not computable,
we can imagine an oracle answering questions about its values.

Given the training set (\ref{eq:training-set}) and the test object $x_{l+1}$,
we can act as follows:
\begin{itemize}
\item
  consider all possible values $Y\in\mathbf{Y}$
  for the label $y_{l+1}$;
\item
  find the randomness level detected by $t$ for every possible completion (\ref{eq:completion});
\item
  predict the label $Y$ corresponding to a completion
  with the largest randomness level detected by $t$;
\item
  output as the \emph{confidence} in this prediction
  one minus the second largest randomness level detected by $t$;
\item
  output as the \emph{credibility} of this prediction
  the randomness level detected by $t$
  of the output prediction $Y$
  (i.e., the largest randomness level detected by $t$ over all possible labels).
\end{itemize}
To understand the intuition behind confidence,
let us tentatively choose a conventional ``significance level'', such as $1\%$.
(In the terminology of this paper, this corresponds to a ``confidence level'' of $99\%$,
i.e.,
$100\%$ minus $1\%$.)
If the confidence in our prediction is $99\%$ or more
and the prediction is wrong,
the actual data sequence belongs to an \emph{a priori} chosen
set of probability at most $1\%$
(the set of all data sequences with randomness level detected by $t$
not exceeding $1\%$).

Intuitively, low credibility means that
either the training set is non-random
or the test object is not representative of the training set
(say, in the training set we have images of digits
and the test object is that of a letter).

\subsection*{Confidence Predictors}

In regression problems,
confidence, as defined in the previous subsection,
is not a useful quantity:
it will typically be equal to 0.
A better approach is to choose a range of confidence levels $1-\epsilon$,
and for each of them specify a \emph{prediction set}
$\Gamma^{\epsilon}\subseteq\mathbf{Y}$,
the set of labels deemed possible at the confidence level $1-\epsilon$.
We will always consider nested prediction sets:
$\Gamma^{\epsilon_1}\subseteq\Gamma^{\epsilon_2}$ when $\epsilon_1\ge\epsilon_2$.
A \emph{confidence predictor} is a function
that maps each training set, each new object, and each confidence level $1-\epsilon$
(formally, we allow $\epsilon$ to take any value in $(0,1)$)
to the corresponding prediction set $\Gamma^{\epsilon}$.
For the confidence predictor to be \emph{valid} the probability that the true label
will fall outside the prediction set $\Gamma^{\epsilon}$ should not exceed $\epsilon$,
for each $\epsilon$.

\begin{figure}
  \centering
  \makebox{\includegraphics[width=\picturewidth,clip=true]{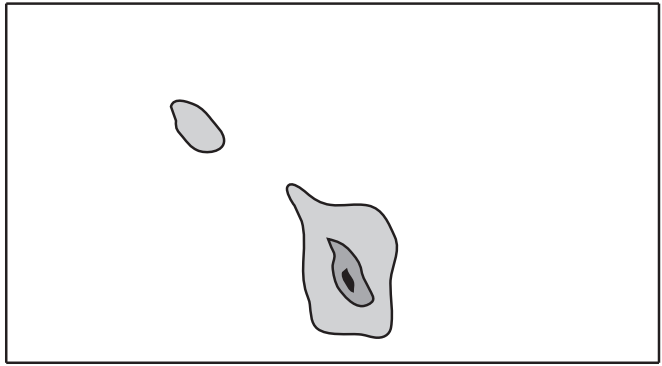}}
  \caption{\label{fig:predset}An example of a nested family of prediction sets
    (casual prediction in black,
    confident prediction in dark grey,
    and highly confident prediction in light grey).}
\end{figure}

We might, for example, choose the confidence levels 99\%, 95\% and 80\%,
and refer to the 99\% prediction set $\Gamma^{1\%}$ as the highly confident prediction,
to the 95\% prediction set $\Gamma^{5\%}$ as the confident prediction,
and to the 80\% prediction set $\Gamma^{20\%}$ as the casual prediction.
Figure \ref{fig:predset} shows how such a family of prediction sets might look
in the case of a rectangular label space $\mathbf{Y}$.
The casual prediction pinpoints the target quite well,
but we know that this kind of prediction can be wrong with probability 20\%.
The confident prediction is much bigger.
If we want to be highly confident
(make a mistake only with probability 1\%),
we must accept an even lower accuracy;
there is even a completely different location that we cannot rule out
at this level of confidence.

Given a randomness test, again universal or not,
we can define the corresponding confidence predictor as follows:
for any confidence level $1-\epsilon$,
the corresponding prediction set consists of the $Y$s
such that the randomness level of the completion (\ref{eq:completion})
detected by the test is greater than $\epsilon$.
The condition (\ref{eq:test-validity}) of validity for statistical tests
implies that a confidence predictor defined in this way
is always valid.

The confidence predictor based on the universal test
(the \emph{universal confidence predictor})
is an interesting object for mathematical investigation
(see, e.g., \cite{vovk/etal:1999}, Section 4),
but it is not computable and so cannot be used in practice.
Our goal in the following sections will be
to find computable approximations to it.

\section{Conformal Prediction}
\label{sec:conformal}


In the previous section we explained how randomness tests
can be used for prediction.
The connection between testing and prediction is, of course, well understood
and have been discussed at length by philosophers \cite{popper:1934}
and statisticians
(see, e.g., the textbook \cite{cox/hinkley:1974}, Section 7.5).
In this section we will see how some popular prediction algorithms
can be transformed into randomness tests
and, therefore, be used for producing hedged predictions.

Let us start with the most successful recent development in machine learning,
support vector machines
(\cite{vapnik:1995,vapnik:1998},
with a key idea going back
to the generalized portrait method \cite{vapnik/chervonenkis:1974}).
Suppose the label space is $\mathbf{Y}=\{-1,1\}$
(we are dealing with the binary classification problem).
With each set of examples
\begin{equation}\label{eq:set}
  (x_1,y_1),
  \ldots,
  (x_n,y_n)
\end{equation}
one associates an optimization problem
whose solution produces nonnegative numbers $\alpha_1,\ldots,\alpha_n$
(``Lagrange multipliers'').
These numbers determine the prediction rule used by the support vector machine
(see \cite{vapnik:1998}, Chapter 10, for details),
but they also are interesting objects in their own right.
Each $\alpha_i$, $i=1,\ldots,n$, tells us
how ``strange'' an element of the set (\ref{eq:set})
the corresponding example $(x_i,y_i)$ is.
If $\alpha_i=0$, $(x_i,y_i)$ fits (\ref{eq:set}) very well
(in fact so well that such examples are uninformative,
and the support vector machine ignores them when making predictions).
The elements with $\alpha_i>0$ are called \emph{support vectors},
and the large value of $\alpha_i$ indicates
that the corresponding $(x_i,y_i)$ is an outlier.

Taking the completion (\ref{eq:completion}) as (\ref{eq:set})
(so that $n=l+1$),
we can find the corresponding $\alpha_1,\ldots,\alpha_{l+1}$.
If $Y$ is different from the actual label $y_{l+1}$,
we expect $(x_{l+1},Y)$ to be an outlier in (\ref{eq:completion})
and so $\alpha_{l+1}$ be large as compared with $\alpha_1,\ldots,\alpha_l$.
A natural way to compare $\alpha_{l+1}$ to the other $\alpha$s
is to look at the ratio
\begin{equation}\label{eq:p}
  p_Y
  :=
  \frac
  {
    \left|
      \{i=1,\ldots,l+1 \st \alpha_i\ge\alpha_{l+1}\}
    \right|
  }
  {l+1},
\end{equation}
which we call the \emph{p-value} associated with the possible label $Y$ for $x_{l+1}$.
In words, the p-value is the proportion of the $\alpha$s
which are at least as large as the last $\alpha$.

The methodology of support vector machines
(as described in \cite{vapnik:1995,vapnik:1998})
is directly applicable
only to the binary classification problems,
but the general case can be reduced to the binary case
by the standard ``one-against-one'' or ``one-against-the-rest'' procedures.
This allows us to define the strangeness values $\alpha_1,\ldots,\alpha_{l+1}$
for general classification problems
(see \cite{vovk/etal:2005}, p.~59, for details),
which in turn determine the p-values (\ref{eq:p}).

The function that assigns to each sequence (\ref{eq:completion})
the corresponding p-value, defined by (\ref{eq:p}),
is a randomness test
(this will follow from Theorem \ref{thm:on-line}
stated in Section \ref{sec:on-line} below).
Therefore, the p-values,
which are our approximations to the corresponding randomness levels,
can be used for hedged prediction
as described in the previous section.
For example, if the p-value $p_{-1}$ is small while $p_1$ is not small,
we can predict $1$ with confidence $1-p_{-1}$ and credibility $p_1$.
Typical credibility will be 1:
for most data sets the percentage of support vectors is small
(\cite{vapnik:1998}, Chapter 12),
and so we can expect $\alpha_{l+1}=0$ when $Y=y_{l+1}$.

\begin{remark*}
  When the order of examples is irrelevant,
  we refer to the data set (\ref{eq:set}) as a set,
  although as a mathematical object it is a multiset rather than a set
  since it can contain several copies of the same example.
  We will continue to use this informal terminology
  (to be completely accurate,
  we would have to say ``data multiset'' instead of ``data set''!)
\end{remark*}



\ifJOURNAL
\begin{table*}
\processtable{Selected test examples from the USPS data set:
  the p-values of digits (0--9), true and predicted labels,
  and confidence and credibility values.\label{tab:examples}}
{\footnotesize\begin{tabular}{|c|c|c|c|c|c|c|c|c|c|c|c|c|c|}
\hline 0 & 1 & 2 & 3 & 4 & 5 & 6 & 7 & 8 & 9 &
  \vbox{\hbox{\strut true}\hbox{\strut label}} &
  \vbox{\hbox{\strut pre-}\hbox{\strut diction}} &
  \vbox{\hbox{\strut confi-}\hbox{\strut dence}} &
  \vbox{\hbox{\strut credi-}\hbox{\strut bility}}\\
\hline 0.01\% & 0.11\% & 0.01\% & 0.01\% & 0.07\% & 0.01\% & 100\% & 0.01\% & 0.01\% & 0.01\%
   & 6 & 6 & 99.89\% & 100\%\\
\hline 0.32\% & 0.38\% & 1.07\% & 0.67\% & 1.43\% & 0.67\% & 0.38\% & 0.33\% & 0.73\% & 0.78\%
   & 6 & 4 & 98.93\% & 1.43\%\\
\hline 0.01\% & 0.27\% & 0.03\% & 0.04\% & 0.18\% & 0.01\% & 0.04\% & 0.01\% & 0.12\% & 100\%
 & 9 & 9 & 99.73\% & 100\%\\
\hline
\end{tabular}}{}
\end{table*}
\fi

\ifnotJOURNAL
\begin{table*}
\caption{Selected test examples from the USPS data set:
  the p-values of digits (0--9), true and predicted labels,
  and confidence and credibility values.\label{tab:examples}}

\medskip

{\tiny\hspace{-12mm}\begin{tabular}{|c|c|c|c|c|c|c|c|c|c|c|c|c|c|}
\hline 0 & 1 & 2 & 3 & 4 & 5 & 6 & 7 & 8 & 9 &
  \vbox{\hbox{\strut true}\hbox{\strut label}} &
  \vbox{\hbox{\strut pre-}\hbox{\strut diction}} &
  \vbox{\hbox{\strut confi-}\hbox{\strut dence}} &
  \vbox{\hbox{\strut credi-}\hbox{\strut bility}}\\
\hline 0.01\% & 0.11\% & 0.01\% & 0.01\% & 0.07\% & 0.01\% & 100\% & 0.01\% & 0.01\% & 0.01\%
   & 6 & 6 & 99.89\% & 100\%\\
\hline 0.32\% & 0.38\% & 1.07\% & 0.67\% & 1.43\% & 0.67\% & 0.38\% & 0.33\% & 0.73\% & 0.78\%
   & 6 & 4 & 98.93\% & 1.43\%\\
\hline 0.01\% & 0.27\% & 0.03\% & 0.04\% & 0.18\% & 0.01\% & 0.04\% & 0.01\% & 0.12\% & 100\%
 & 9 & 9 & 99.73\% & 100\%\\
\hline
\end{tabular}}{}
\end{table*}
\fi

Table~\ref{tab:examples} illustrates the results of hedged prediction
for a popular data set of hand-written digits
called the USPS data set \cite{lecun/etal:1990}.
The data set contains 9298 digits represented as a $16\times16$ matrix of pixels;
it is divided into a training set of size 7291 and a test set of size 2007.
For several test examples the table shows
the p-values for each possible label, the actual label,
the predicted label, confidence, and credibility,
computed using the support vector method with the polynomial kernel of degree 5.
To interpret the numbers in this table,
remember that high (i.e., close to 100\%) confidence
means that all labels except the predicted one are unlikely.
If, say, the first example were predicted wrongly,
this would mean that a rare event (of probability less than 1\%) had occurred;
therefore, we expect the prediction to be correct (which it is).
In the case of the second example,
confidence is also quite high (more than 95\%),
but we can see that the credibility is low (less than 5\%).
From the confidence we can conclude that the labels other than 4
are excluded at level 5\%,
but the label 4 itself is also excluded at the level 5\%.
This shows that the prediction algorithm
was unable to extract from the training set enough information
to allow us to confidently classify this example:
the strangeness of the labels different from 4 may be due
to the fact that the object itself is strange;
perhaps the test example is very different from all examples in the training set.
Unsurprisingly, the prediction for the second example is wrong.

In general,
high confidence shows that all alternatives
to the predicted label are unlikely.
Low credibility means that the whole situation is suspect;
as we have already mentioned, we will obtain a very low credibility
if the new example is a letter (whereas all training examples are digits).
Credibility will also be low if the new example is a digit
written in an unusual way.
Notice that typically credibility will not be low
provided the data set was generated independently from the same distribution:
the probability that credibility
will not exceed some threshold $\epsilon$ (such as 1\%)
is at most $\epsilon$.
In summary,
we can trust a prediction if
(1) the confidence is close to 100\% and
(2) the credibility is not low (say, is not less than 5\%).

Many other prediction algorithms can be used as underlying algorithms
for hedged prediction.
For example, we can use the nearest neighbours technique to associate
\begin{equation}\label{eq:NN}
  \alpha_i
  :=
  \frac
  {\sum_{j=1}^k d_{ij}^+}
  {\sum_{j=1}^k d_{ij}^-},
  \quad
  i=1,\ldots,n,
\end{equation}
with the elements $(x_i,y_i)$ of the set (\ref{eq:set}),
where $d_{ij}^+$ is the $j$th shortest distance from $x_i$
to other objects labelled in the same way as $x_i$,
and $d_{ij}^-$ is the $j$th shortest distance
from $x_i$ to the objects labelled differently from $x_i$;
the parameter $k\in\{1,2,\dots\}$ in~(\ref{eq:NN})
is the number of nearest neighbours taken into account.
The distances can be computed in a feature space
(that is, the distance between $x\in\mathbf{X}$ and $x'\in\mathbf{X}$
can be understood as $\left\|F(x)-F(x')\right\|$,
$F$ mapping the object space $\mathbf{X}$ into a feature, typically Hilbert, space),
and so (\ref{eq:NN}) can also be used with the kernel nearest neighbours.

The intuition behind (\ref{eq:NN}) is as follows:
a typical object $x_i$ labelled by, say, $y$
will tend to be surrounded by other objects labelled by $y$;
and if this is the case, the corresponding $\alpha_i$ will be small.
In the untypical case that there are objects whose labels are different from $y$
nearer than objects labelled $y$,
$\alpha_i$ will become larger.
Therefore, the $\alpha$s reflect the strangeness of examples.

The p-values computed by (\ref{eq:NN})
can again be used for hedged prediction.
It is a general empirical fact that
the accuracy and reliability of the hedged predictions
are in line with the error rate of the underlying algorithm.
For example, in the case of the USPS data set,
the 1-nearest neighbour algorithm
(i.e., the one with $k=1$)
achieves the error rate of 2.2\%,
and the hedged predictions based on (\ref{eq:NN}) are highly confident
(achieve confidence of at least $99\%$)
for more than 95\% of the test examples.

\subsection*{General Definition}

The general notion of conformal predictor can be defined as follows.
A \emph{nonconformity measure} is a function that assigns
to every data sequence (\ref{eq:set}) a sequence of numbers
$\alpha_1,\ldots,\alpha_n$,
called \emph{nonconformity scores},
in such a way that interchanging any two examples $(x_i,y_i)$ and $(x_j,y_j)$
leads to the interchange of the corresponding nonconformity scores $\alpha_i$ and $\alpha_j$
(with all the other nonconformity scores unaffected).
The corresponding \emph{conformal predictor} maps each data set (\ref{eq:training-set}),
$l=0,1,\ldots$,
each new object $x_{l+1}$,
and each confidence level $1-\epsilon\in(0,1)$,
to the prediction set
\begin{equation}\label{eq:Gamma}
  \Gamma^{\epsilon}
  \left(
    x_1,y_1,\ldots,x_{l},y_{l},x_{l+1}
  \right)
  :=
  \left\{
    Y\in\mathbf{Y}
    \st
    p_Y
    >
    \epsilon
  \right\},
\end{equation}
where $p_Y$ are defined by (\ref{eq:p})
with $\alpha_1,\ldots,\alpha_{l+1}$ being the nonconformity scores
corresponding to the data sequence (\ref{eq:completion}).

We have already remarked that associating with each completion (\ref{eq:completion})
the p-value (\ref{eq:p}) gives a randomness test;
this is true in general.
This implies that for each $l$ the probability of the event
\begin{equation*}
  y_{l+1}
  \in
  \Gamma^{\epsilon}
  \left(
    x_1,y_1,\ldots,x_{l},y_{l},x_{l+1}
  \right)
\end{equation*}
is at least $1-\epsilon$.

This definition works for both classification and regression,
but in the case of classification we can summarize (\ref{eq:Gamma})
by two numbers:
the confidence
\begin{equation}\label{eq:conf}
  \sup
  \left\{
    1-\epsilon
    \st
    \left|
      \Gamma^{\epsilon}
    \right|
    \le
    1
  \right\}
\end{equation}
and the credibility
\begin{equation}\label{eq:cred}
  \inf
  \left\{
    \epsilon
    \st
    \left|
      \Gamma^{\epsilon}
    \right|
    =
    0
  \right\}.
\end{equation}

\subsection*{Computationally Efficient Regression}

As we have already mentioned,
the algorithms described so far
cannot be applied directly in the case of regression,
even if the randomness test is efficiently computable:
now we cannot consider all possible values $Y$ for $y_{l+1}$
since there are infinitely many of them.
However, there might still be computationally efficient
ways to find the prediction sets $\Gamma^{\epsilon}$.
The idea is that if $\alpha_i$ are defined as the residuals
\begin{equation}\label{eq:residual}
  \alpha_i
  :=
  \left|
    y_i - f_Y(x_i)
  \right|
\end{equation}
where $f_Y:\mathbf{X}\to\bbbr$ is a regression function
fitted to the completed data set~(\ref{eq:completion}),
then $\alpha_i$ may have a simple expression in terms of $Y$,
leading to an efficient way of computing the prediction sets
(via (\ref{eq:p}) and (\ref{eq:Gamma})).
This idea was implemented in \cite{nouretdinov/etal:2001rr}
in the case where $f_Y$ is found from the ridge regression,
or kernel ridge regression, procedure,
with the resulting algorithm of hedged prediction
called the \emph{ridge regression confidence machine}.
For a much fuller description of the ridge regression confidence machine
(and its modifications in the case where (\ref{eq:residual})
are replaced by the fancier ``deleted'' or ``studentized'' residuals)
see \cite{vovk/etal:2005}, Section 2.3.

\section{Bayesian Approach to Conformal Prediction}
\label{sec:Bayesian}

Bayesian methods have become very popular in both machine learning and statistics
thanks to their power and versatility,
and in this section we will see
how Bayesian ideas can be used for designing efficient conformal predictors.
We will only describe results of computer experiments
(following \cite{melluish/etal:2001})
with artificial data sets,
since for real-world data sets there is no way
to make sure that the Bayesian assumption is satisfied.

Suppose $\mathbf{X}=\bbbr^p$
(each object is a vector of $p$ real-valued attributes)
and our model of the data-generating mechanism is
\begin{equation}\label{eq:model}
  y_i
  =
  w\cdot x_i
  +
  \xi_i,
  \quad
  i=1,2,\ldots,
\end{equation}
where $\xi_i$ are independent standard Gaussian random variables
and the weight vector $w\in\bbbr^p$ is distributed as $N(0,(1/a)I_p)$
(we use the notation $I_p$ for the unit $p\times p$ matrix
and $N(0,A)$ for the $p$-dimensional Gaussian distribution
with covariance matrix $A$);
$a$ is a positive constant.
The actual data-generating mechanism used in our experiments
will correspond to this model with $a$ set to 1.

Under the model (\ref{eq:model}) the best (in the mean-square sense) fit
to a data set (\ref{eq:set})
is provided by the ridge regression procedure with parameter $a$
(for details, see, e.g., \cite{vovk/etal:2005}, Section 10.3).
Using the residuals (\ref{eq:residual}) with $f_Y$
found by ridge regression with parameter $a$
leads to an efficient conformal predictor
which will be referred to as the ridge regression confidence machine with parameter $a$.
Each prediction set output by the ridge regression confidence machine
will be replaced by its convex hull,
the corresponding \emph{prediction interval}.

\begin{figure}
  \centering
  \makebox{\includegraphics[width=\picturewidth,clip=true]{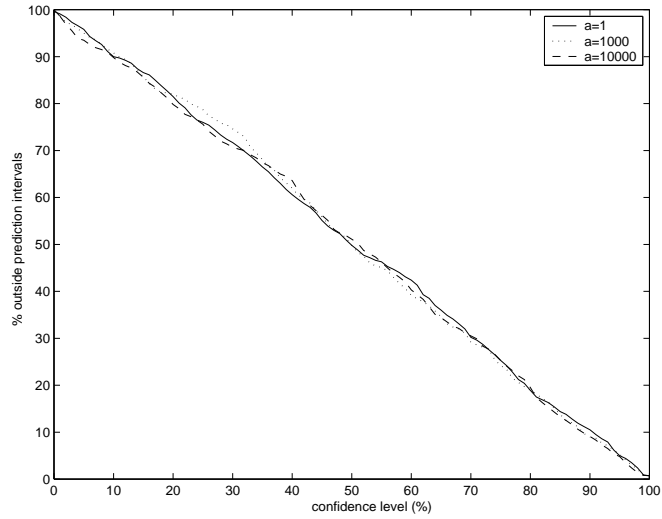}}
  \caption{\label{fig:rrcm-errors}Validity for the ridge regression confidence machine.}
\end{figure}

To test the validity and efficiency of the ridge regression confidence machine
the following procedure was used.
Ten times a vector $w\in\bbbr^5$ was independently generated from the distribution $N(0,I_5)$.
For each of the 10 values of $w$,
100 training objects and 100 test objects
were independently generated from the uniform distribution on $[-10,10]^5$
and for each object $x$ its label $y$ was generated as $w\cdot x+\xi$,
with all the $\xi$ standard Gaussian and independent.
For each of the 1000 test objects and each confidence level $1-\epsilon$
the prediction set $\Gamma^{\epsilon}$ for its label
was found from the corresponding training set
using the ridge regression confidence machine with parameter $a=1$.
The solid line in Figure~\ref{fig:rrcm-errors} shows the confidence level
against the percentage of test examples whose labels
were not covered by the corresponding prediction intervals at that confidence level.
Since conformal predictors are always valid,
the percentage outside the prediction interval
should never exceed 100 minus the confidence level,
up to statistical fluctuations,
and this is confirmed by the picture.

\begin{figure}
  \centering
  \makebox{\includegraphics[width=\picturewidth,clip=true]{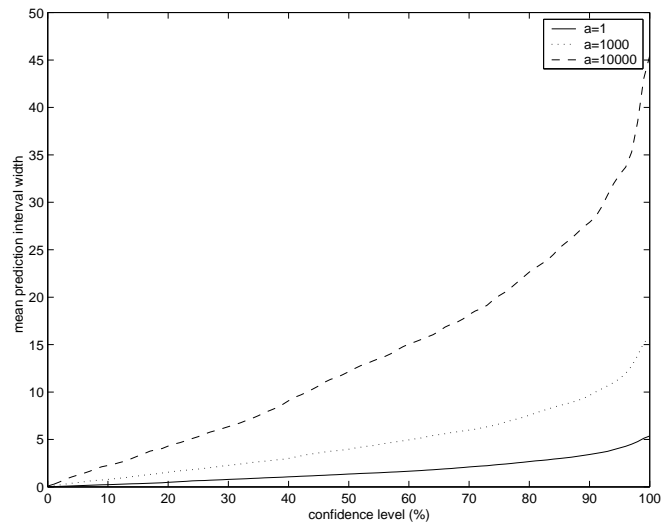}}
  \caption{\label{fig:rrcm-widths}Efficiency for the ridge regression confidence machine.}
\end{figure}

A natural measure of efficiency of confidence predictors
is the mean width of their prediction intervals,
at different confidence levels:
the algorithm is the more efficient the narrower prediction intervals it produces.
The solid line in Figure~\ref{fig:rrcm-widths} shows
the confidence level against the mean
(over all test examples)
width of the prediction intervals at that confidence level.

\begin{figure}
  \centering
  \makebox{\includegraphics[width=\picturewidth,clip=true]{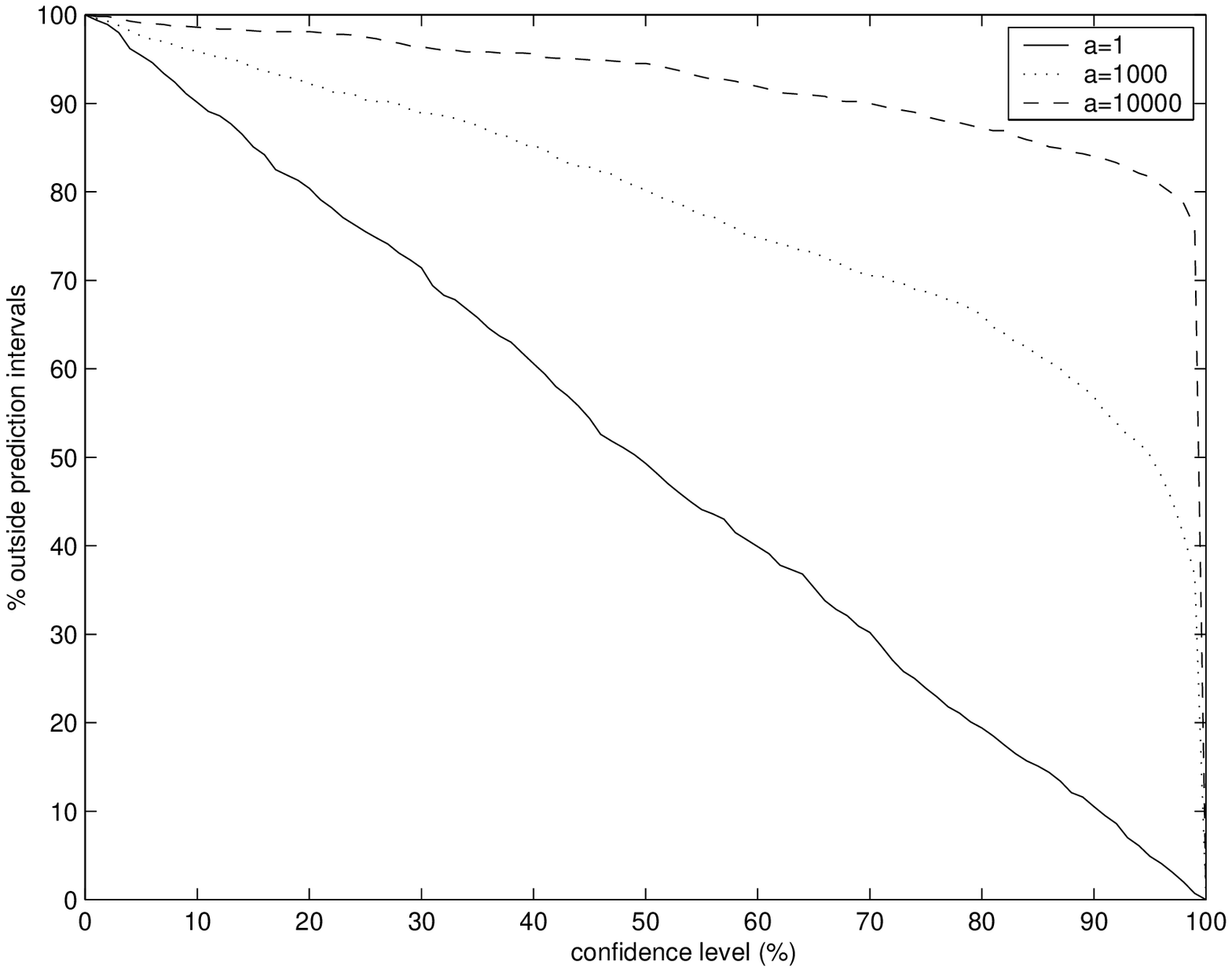}}
  \caption{\label{fig:Bayes-errors}Validity for the Bayes-optimal confidence predictor.}
\end{figure}

\begin{figure}
  \centering
  \makebox{\includegraphics[width=\picturewidth,clip=true]{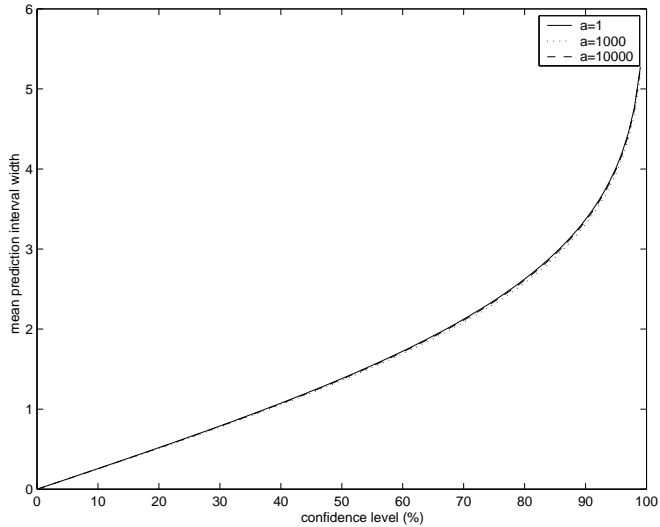}}
  \caption{\label{fig:Bayes-widths}Efficiency for the Bayes-optimal confidence predictor.}
\end{figure}

Since we know the data-generating mechanism,
the approach via conformal prediction appears somewhat roundabout:
for each test object we could instead find
the conditional probability distribution of its label,
which is Gaussian,
and output as the prediction set $\Gamma^{\epsilon}$
the shortest 
(i.e., centred at the mean of the conditional distribution)
interval of conditional probability $1-\epsilon$.
Figures \ref{fig:Bayes-errors} and \ref{fig:Bayes-widths}
are the analogues of Figures \ref{fig:rrcm-errors} and \ref{fig:rrcm-widths}
for this \emph{Bayes-optimal confidence predictor}.
The solid line in Figure \ref{fig:Bayes-errors}
demonstrates the validity of the Bayes-optimal confidence predictor.

What is interesting is that the solid lines
in Figures~\ref{fig:Bayes-widths} and \ref{fig:rrcm-widths}
look exactly the same,
taking account of the different scales of the vertical axes.
The ridge regression confidence machine
appears as good as the Bayes-optimal predictor.
(This is a general phenomenon;
it is also illustrated, in the case of classification,
by the construction in Section 3.3 of \cite{vovk/etal:2005}
of a conformal predictor that is asymptotically
as good as the Bayes-optimal confidence predictor.)

The similarity between the two algorithms disappears
when they are given wrong values for $a$.
For example,
let us see what happens if we tell the algorithms
that the expected value of $\|w\|$ is just $1\%$ of what it really is
(this corresponds to taking $a=10000$).
The ridge regression confidence machine stays valid
(see the dashed line in Figure \ref{fig:rrcm-errors}),
but its efficiency deteriorates
(the dashed line in Figure \ref{fig:rrcm-widths}).
The efficiency of the Bayes-optimal confidence predictor
(the dashed line in Figure \ref{fig:Bayes-widths})
is hardly affected,
but its predictions become invalid
(the dashed line in Figure \ref{fig:Bayes-errors}
deviates significantly from the diagonal,
especially for the most important large confidence levels:
e.g., only about 15\% of labels fall within the 90\% prediction sets).
The worst that can happen to the ridge regression confidence machine
is that its predictions will become useless
(but at least harmless),
whereas the Bayes-optimal predictions can become misleading.

Figures \ref{fig:rrcm-errors}--\ref{fig:Bayes-widths} also show the graphs
for the intermediate value $a=1000$.
Similar results but for different data sets
are also given in \cite{vovk/etal:2005}, Section 10.3.
A general scheme of Bayes-type conformal prediction
is described in \cite{vovk/etal:2005}, pp.~102--103.

\section{On-line prediction}
\label{sec:on-line}


We know from Section \ref{sec:conformal}
that conformal predictors are valid in the sense that the probability of error
\begin{equation}\label{eq:error}
  y_{l+1}
  \notin
  \Gamma^{\epsilon}
  \left(
    x_1,y_1,
    \ldots
    x_l,y_l,
    x_{l+1}
  \right)
\end{equation}
at confidence level $1-\epsilon$
never exceeds $\epsilon$.
The word ``probability'' means ``unconditional probability'' here:
the frequentist meaning of the statement that the probability of (\ref{eq:error})
does not exceed $\epsilon$
is that,
if we repeatedly generate many sequences
\begin{equation*}
  x_1,y_1,\ldots,x_l,y_l,x_{l+1},y_{l+1},
\end{equation*}
the fraction of them satisfying (\ref{eq:error})
will be at most $\epsilon$,
to within statistical fluctuations.
To say that we are controlling the number of errors
would be an exaggeration
because of the artificial character of this scheme
of repeatedly generating a new training set and a new test example.
Can we say that the confidence level $1-\epsilon$
translates into a bound on the number of mistakes
for a natural learning protocol?
In this section we show that the answer is ``yes''
for the popular on-line learning protocol,
and in the next section we will see to what degree
this carries over to other protocols.

In on-line learning the examples are presented one by one.
Each time, we observe the object and predict its label.
Then we observe the label and go on to the next example.
We start by observing the first object $x_1$ and predicting its label $y_1$.
Then we observe $y_1$ and the second object $x_2$, and predict its label $y_2$.
And so on.
At the $n$th step,
we have observed the previous examples
$ 
  (x_1,y_1),\dots,(x_{n-1},y_{n-1})
$ 
and the new object $x_n$, and our task is to predict $y_n$.
The quality of our predictions should improve
as we accumulate more and more old examples.
This is the sense in which we are learning.

Our prediction for $y_n$ is a nested family of prediction sets
$\Gamma_n^{\epsilon}\subseteq\mathbf{Y}$,
$\epsilon\in(0,1)$.
The process of prediction can be summarized by the following protocol:

\medskip

\noindent\textsc{On-line prediction protocol}
\ifJOURNAL
  \newcommand{\Indent}{\quad}
\fi
\ifnotJOURNAL
  \newcommand{\Indent}{\quad\enspace}

  \smallskip

\fi

\noindent
\Indent$\Err_0:=0$;

\noindent
\Indent$\Mult_0:=0$;

\noindent
\Indent$\Emp_0:=0$;

\noindent
\Indent FOR $n=1,2,\ldots$:

\noindent
\Indent\Indent Reality outputs $x_n\in\mathbf{X}$;

\noindent
\Indent\Indent Predictor outputs $\Gamma_n^{\epsilon}\subseteq\mathbf{Y}$ for all $\epsilon\in(0,1)$;

\noindent
\Indent\Indent Reality outputs $y_n\in\mathbf{Y}$;

\noindent
\Indent\Indent$\err_n^{\epsilon}
  :=
  \left\{
    \begin{array}{ll}
      1 & \text{if $y_n \notin \Gamma_n^{\epsilon}$}\\
      0 & \text{otherwise},
    \end{array}
  \right.
  \quad
  \epsilon\in(0,1)$;

\noindent
\Indent\Indent\strut$\Err_n^{\epsilon}:=\Err^{\epsilon}_{n-1}+\err_n^{\epsilon},
  \quad
  \epsilon\in(0,1)$;

\noindent
\Indent\Indent$\mult_n^{\epsilon}
  :=
  \left\{
    \begin{array}{ll}
      1 & \text{if $\left|\Gamma_n^{\epsilon}\right|>1$}\\
      0 & \text{otherwise},
    \end{array}
  \right.
  \quad
  \epsilon\in(0,1)$;

\noindent
\Indent\Indent\strut$\Mult_n^{\epsilon}:=\Mult_{n-1}^{\epsilon}+\mult_n^{\epsilon},
  \quad
  \epsilon\in(0,1)$;

\noindent
\Indent\Indent$\emp_n^{\epsilon}
  :=
  \left\{
    \begin{array}{ll}
      1 & \text{if $\left|\Gamma_n^{\epsilon}\right|=0$}\\
      0 & \text{otherwise},
    \end{array}
  \right.
  \quad
  \epsilon\in(0,1)$;

\noindent
\Indent\Indent\strut$\Emp_n^{\epsilon}:=\Emp_{n-1}^{\epsilon}+\Emp_n^{\epsilon},
  \quad
  \epsilon\in(0,1)$

\noindent
\Indent END FOR.

\medskip

\noindent
As we said, the family $\Gamma_n^{\epsilon}$
is assumed nested:
$\Gamma_n^{\epsilon_1}\subseteq\Gamma_n^{\epsilon_2}$ when $\epsilon_1\ge\epsilon_2$.
In this protocol we also record the cumulative numbers
$\Err_n^{\epsilon}$ of erroneous prediction sets,
$\Mult_n^{\epsilon}$ of \emph{multiple} prediction sets
(i.e., prediction sets containing more than one label)
and $\Emp_n^{\epsilon}$ of empty prediction sets
at each confidence level $1-\epsilon$.
We will discuss the significance of each of these numbers in turn.

\begin{figure}
  \centering
  \makebox{\includegraphics[width=\picturewidth]{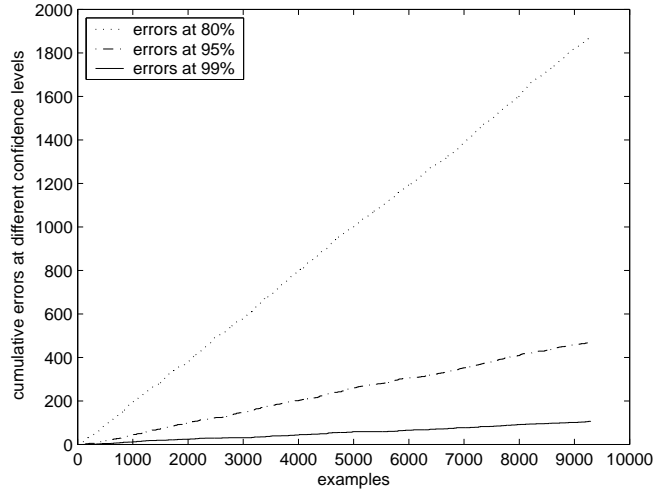}}
  \caption{\label{fig:CP0err}Cumulative numbers of errors for a conformal predictor
    (the 1-nearest neighbour conformal predictor)
    run in the on-line mode on the USPS data set
    (9298 hand-written digits, randomly permuted)
    at the confidence levels 80\%, 95\% and 99\%.}
\end{figure}

The number of erroneous predictions is a measure of validity of our confidence predictors:
we would like to have $\Err_n^{\epsilon}\le\epsilon n$,
up to statistical fluctuations.
In Figure~\ref{fig:CP0err} we can see the lines $n\mapsto\Err_n^{\epsilon}$
for one particular conformal predictor
and for three confidence levels $1-\epsilon$:
the solid line for 99\%, the dash-dot line for 95\%, and the dotted line for 80\%.
The number of errors made grows linearly,
and the slope is approximately
20\% for the confidence level 80\%,
5\% for the confidence level 95\%,
and 1\% for the confidence level 99\%.
We will see below that this is not accidental.

\begin{figure}
  \centering
  \makebox{\includegraphics[width=\picturewidth]{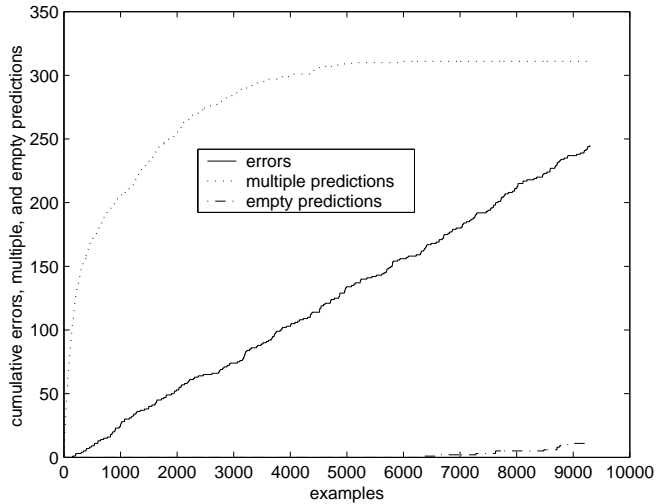}}
  \caption{\label{fig:TCM975}The on-line performance of the 1-nearest neighbour conformal predictor
    at the confidence level 97.5\% on the USPS data set (randomly permuted).}
\end{figure}

The number of multiple predictions $\Mult_n$
is a useful measure of efficiency in the case of classification:
we would like as many as possible of our predictions to be singletons.
Figure \ref{fig:TCM975} shows the cumulative numbers of errors
$n\mapsto\Err_n^{2.5\%}$ (solid line)
and multiple predictions
$n\mapsto\Mult_n^{2.5\%}$ (dotted line)
at the fixed confidence level 97.5\%.
We can see that out of approximately 10,000 predictions
about 250 (approximately 2.5\%) were errors
and about 300 (approximately 3\%) were multiple predictions.

We can see that by choosing $\epsilon$ we are able to control the number of errors.
For small $\epsilon$
(relative to the difficulty of the data set)
this might lead to the need sometimes to give
multiple predictions.
On the other hand,
for larger $\epsilon$ this might lead to empty predictions at some steps,
as can be seen from the bottom right corner of Figure \ref{fig:TCM975}:
when the predictor ceases to make multiple predictions
it starts making occasional empty predictions
(the dash-dot line).
An empty prediction is a warning that the object to be predicted is unusual
(the credibility, as defined in Section \ref{sec:ideal}, is $\epsilon$ or less).

It would be a mistake to concentrate exclusively on one confidence level $1-\epsilon$.
If the prediction $\Gamma_n^{\epsilon}$ is empty,
this does not mean that we cannot make any prediction at all:
we should just shift our attention to other confidence levels
(perhaps look at the range of $\epsilon$ for which $\Gamma_n^{\epsilon}$ is a singleton).
Likewise, $\Gamma_n^{\epsilon}$ being multiple
does not mean that all labels in $\Gamma_n^{\epsilon}$ are equally likely:
slightly increasing $\epsilon$ might lead to the removal of some labels.
Of course,
taking in the continuum of predictions sets, for all $\epsilon\in(0,1)$,
might be too difficult or tiresome for a human mind,
and concentrating on a few conventional levels,
as in Figure \ref{fig:predset},
might be a reasonable compromise.

\ifJOURNAL
\begin{table*}
\processtable{A selected test example from a data set of hospital records of patients
  who suffered acute abdominal pain \cite{gammerman/thatcher:1992}:
  the p-values for the nine possible diagnostic groups
  (appendicitis APP, diverticulitis DIV, perforated peptic ulcer PPU,
  non-specific abdominal pain NAP, cholecystitis CHO, intestinal obstruction INO,
  pancreatitis PAN, renal colic RCO, dyspepsia DYS)
  and the true label.\label{tab:abdominal}}
{\footnotesize\begin{tabular}{|c|c|c|c|c|c|c|c|c|c|}
\hline APP & DIV & PPU & NAP & CHO & INO & PAN & RCO & DYS & true label\\
\hline 1.23\% & 0.36\% & 0.16\% & 2.83\% & 5.72\% & 0.89\% & 1.37\% & 0.48\% & 80.56\% & DYS\\
\hline
\end{tabular}}{}
\end{table*}
\fi

\ifnotJOURNAL
\begin{table*}
\caption{A selected test example from a data set of hospital records of patients
  who suffered acute abdominal pain \cite{gammerman/thatcher:1992}:
  the p-values for the nine possible diagnostic groups
  (appendicitis APP, diverticulitis DIV, perforated peptic ulcer PPU,
  non-specific abdominal pain NAP, cholecystitis CHO, intestinal obstruction INO,
  pancreatitis PAN, renal colic RCO, dyspepsia DYS)
  and the true label.\label{tab:abdominal}}

\medskip

{\footnotesize\hspace{-2mm}\begin{tabular}{|c|c|c|c|c|c|c|c|c|c|}
\hline APP & DIV & PPU & NAP & CHO & INO & PAN & RCO & DYS & true label\\
\hline 1.23\% & 0.36\% & 0.16\% & 2.83\% & 5.72\% & 0.89\% & 1.37\% & 0.48\% & 80.56\% & DYS\\
\hline
\end{tabular}}{}
\end{table*}
\fi

%
%

For example, Table \ref{tab:abdominal}
gives the p-values for different kinds of abdominal pain
obtained for a specific patient based on his symptoms.
We can see that at the confidence level 95\% the prediction set
is multiple,
$\{$cholecystitis, dyspepsia$\}$.
When we relax the confidence level to 90\%,
the prediction set narrows down to $\{$dyspepsia$\}$
(the singleton containing only the true label);
on the other hand,
at the confidence level 99\% the prediction set widens to
$\{$appendicitis, non-specific abdominal pain, cholecystitis, pancreatitis, dyspepsia$\}$.
Such detailed confidence information,
in combination with the property of validity,
is especially valuable in medicine
(and some of the first applications of conformal predictors
have been to the fields of medicine and bioinformatics:
see, e.g., \cite{bellotti/etal:2005,shahmuradov/etal:2005}).

In the case of regression,
we will usually have $\Mult_n^{\epsilon}=n$ and $\Emp_n^{\epsilon}=0$,
and so these are not useful measures of efficiency.
Better measures,
such as the ones used in the previous section,
would, e.g., take into account the widths of the prediction intervals.

\subsection*{Theoretical Analysis}

Looking at Figures \ref{fig:CP0err} and \ref{fig:TCM975}
we might be tempted to guess that the probability of error
at each step of the on-line protocol
is $\epsilon$
and that errors are made independently at different steps.
This is not literally true,
as a closer examination of the bottom left corner of Figure \ref{fig:TCM975} reveals.
It, however, becomes true
(as noticed in \cite{vovk:2002})
if the p-values (\ref{eq:p}) are redefined as
\begin{equation}\label{eq:p-smoothed}
  p_Y
  :=
  \frac
  {
    \left|
      \{i \st \alpha_i>\alpha_{l+1}\}
    \right|
    +
    \eta
    \left|
      \{i \st \alpha_i=\alpha_{l+1}\}
    \right|
  }
  {l+1},
\end{equation}
where $i$ ranges over $\{1,\ldots,l+1\}$
and $\eta\in[0,1]$ is generated randomly from the uniform distribution on $[0,1]$
(the $\eta$s should be independent between themselves and of everything else;
in practice they are produced by pseudo-random number generators).
The only difference between (\ref{eq:p}) and (\ref{eq:p-smoothed})
is that the expression (\ref{eq:p-smoothed}) takes more care in breaking the ties
$\alpha_i=\alpha_{l+1}$.
Replacing (\ref{eq:p}) by (\ref{eq:p-smoothed})
in the definition of conformal predictor
we obtain the notion of \emph{smoothed conformal predictor}.

The validity property for smoothed conformal predictors can now be stated as follows.
\begin{theorem}\label{thm:on-line}
  Suppose the examples
  \begin{equation*}
    (x_1,y_1),(x_2,y_2),\ldots
  \end{equation*}
  are generated independently
  from the same distribution.
  For any smoothed conformal predictor working in the on-line prediction protocol
  and any confidence level $1-\epsilon$,
  the random variables $\err_1^{\epsilon},\err_2^{\epsilon},\ldots$
  are independent and take value 1 with probability $\epsilon$.
\end{theorem}

Combining Theorem \ref{thm:on-line}
with the strong law of large numbers
we can see that
\begin{equation*}
  \lim_{n\to\infty}
  \frac{\Err_n^{\epsilon}}{n}
  =
  \epsilon
\end{equation*}
holds with probability one for smoothed conformal predictors.
(They are ``well calibrated''.)
Since the number of mistakes made by a conformal predictor
never exceeds the number of mistakes
made by the corresponding smoothed conformal predictor,
\begin{equation*}
  \limsup_{n\to\infty}
  \frac{\Err_n^{\epsilon}}{n}
  \le
  \epsilon
\end{equation*}
holds with probability one for conformal predictors.
(They are ``conservatively well calibrated''.)

\section{Slow teachers, lazy teachers, and the batch setting}
\label{sec:slow}


In the pure on-line setting, considered in the previous section,
we get an immediate feedback (the true label) for every example that we predict.
This makes practical applications of this scenario questionable.
Imagine, for example, a mail sorting centre
using an on-line prediction algorithm
for zip code recognition;
suppose the feedback about the ``true'' label comes from a human ``teacher''.
If the feedback is given for every object $x_i$,
there is no point in having the prediction algorithm:
we can just as well use the label provided by the teacher.
It would help if the prediction algorithm could still work well,
in particular be valid,
if only every, say, tenth object were classified by a human teacher
(the scenario of ``lazy'' teachers).
Alternatively,
even if the prediction algorithm requires the knowledge of all labels,
it might still be useful if the labels were allowed to be given not immediately
but with a delay (``slow'' teachers).
In our mail sorting example,
such a delay might make sure that we hear
from local post offices about any mistakes made
before giving a feedback to the algorithm.

In the pure on-line protocol we had validity in the strongest possible sense:
at each confidence level $1-\epsilon$ each smoothed conformal predictor
made errors independently with probability $\epsilon$.
In the case of weaker teachers
(as usual, we are using the word ``teacher'' in the general sense of the entity
providing the feedback,
called Reality in the previous section),
we have to accept a weaker notion of validity.
Suppose the predictor receives a feedback from the teacher
at the end of steps $n_1,n_2,\ldots$,
$n_1<n_2<\cdots$;
the feedback is the label of one of the objects that the predictor
has already seen (and predicted).
This scheme \cite{ryabko/etal:2003} covers both slow and lazy teachers
(as well as teachers who are both slow and lazy).
It was proved in \cite{nouretdinov/vovk:2003}
(see also \cite{vovk/etal:2005}, Theorem 4.2)
that the smoothed conformal predictors
(using only the examples with known labels)
remain valid in the sense
\begin{equation*}
  \forall\epsilon\in(0,1):
  \Err_n^{\epsilon}/n\to\epsilon
  \text{ in probability}
\end{equation*}
if and only if $n_k/n_{k-1}\to1$ as $k\to\infty$.
In other words,
the validity in the sense of convergence in probability holds
if and only if the growth rate of $n_k$ is subexponential.
(This condition is amply satisfied for our example
of a teacher giving feedback for every tenth object.)

\begin{figure}
  \centering
  \makebox{\includegraphics[width=\picturewidth]{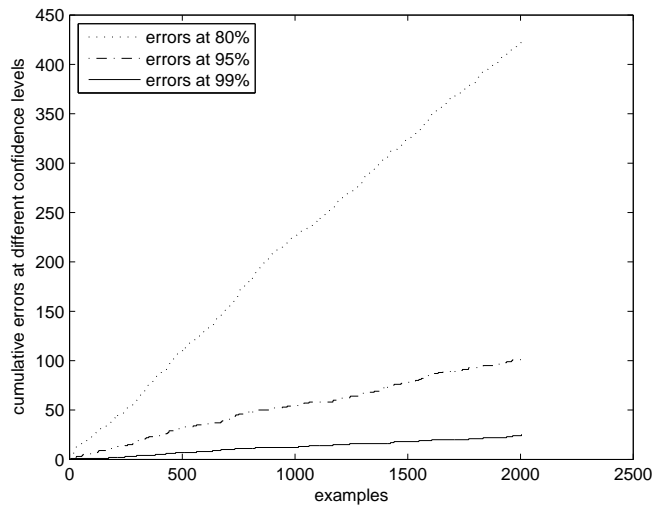}}
  \caption{\label{fig:batch-errors}Cumulative numbers of errors made on the test set
    by the 1-nearest neighbour conformal predictor
    used in the batch mode on the USPS data set
    (randomly permuted and split into a training set of size 7291 and a test set of size 2007)
    at the confidence levels 80\%, 95\% and 99\%.}
\end{figure}

The most standard \emph{batch} setting of the problem of prediction
is in one respect even more demanding than our scenarios of weak teachers.
In this setting we are given a training set (\ref{eq:training-set})
and our goal is to predict the labels
given the objects in the test set
\begin{equation}\label{eq:test-set}
  (x_{l+1},y_{l+1}),\ldots,(x_{l+k},y_{l+k}).
\end{equation}
This can be interpreted as a finite-horizon version
of the lazy-teacher setting:
no labels are returned after step $l$.
Computer experiments (see, e.g., Figure \ref{fig:batch-errors})
show that approximate validity still holds;
for related theoretical results,
see \cite{vovk/etal:2005}, Section 4.4.

\section{Induction and transduction}
\label{sec:induction-transduction}

\begin{figure}
  \centering
  \input{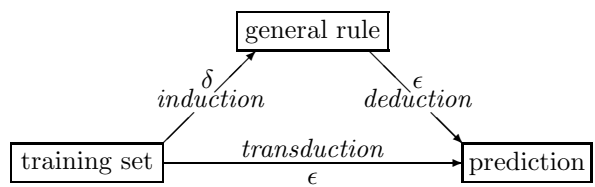}
  \caption{\label{fig:trans}Inductive and transductive prediction.}
\end{figure}

Vapnik's \cite{vapnik:1995,vapnik:1998}
distinction between induction and transduction,
as applied to the problem of prediction,
is depicted in Figure \ref{fig:trans}.
In \emph{inductive prediction}
we first move from examples in hand to some more or less general rule,
which we might call a prediction or decision rule,
a model, or a theory;
this is the \emph{inductive step}.
When presented with a new object,
we derive a prediction from the general rule;
this is the \emph{deductive step}.
In \emph{transductive prediction},
we take a shortcut,
moving from the old examples directly
to the prediction about the new object.

Typical examples of the inductive step
are estimating parameters in statistics
and finding an approximating function
in statistical learning theory.
Examples of transductive prediction
are estimation of future observations in statistics
(\cite{cox/hinkley:1974}, Section 7.5, \cite{takeuchi:1975})
and nearest neighbours algorithms
in machine learning.

In the case of simple (i.e., traditional, not hedged) predictions
the distinction between induction and transduction
is less than crisp.
A method for doing transduction,
in the simplest setting of predicting one label,
is a method for predicting $y_{l+1}$
from (\ref{eq:training-set}) and $x_{l+1}$.
Such a method gives a prediction for any object
that might be presented as $x_{l+1}$, and so it defines,
at least implicitly, a rule,
which might be extracted from the training set (\ref{eq:training-set}) (induction),
stored, and then subsequently applied to $x_{l+1}$ to predict $y_{l+1}$ (deduction).
So any real distinction is really at a practical and computational level:
do we extract and store the general rule or not?

For hedged predictions the difference between induction and transduction goes deeper.
We will typically want different notions of hedged prediction
in the two frameworks.
Mathematical results about induction usually involve two parameters,
often denoted $\epsilon$ (the desired accuracy of the prediction rule)
and $\delta$ (the probability of achieving the accuracy of $\epsilon$),
whereas results about transduction involve only one parameter,
which we denote $\epsilon$ in this paper
(the probability of error we are willing to tolerate);
see Figure \ref{fig:trans}.
For a review of inductive prediction
from this point of view, see \cite{vovk/etal:2005}, Section 10.1.

\section{Inductive conformal predictors}
\label{sec:ICP}


Our approach to prediction is thoroughly transductive,
and this is what makes valid and efficient hedged prediction possible.
In this section we will see, however,
that there is also room for an element of induction
in conformal prediction.

Let us take a closer look at the process of conformal prediction,
as described in Section \ref{sec:conformal}.
Suppose we are given a training set (\ref{eq:training-set})
and the objects in a test set (\ref{eq:test-set}),
and our goal is to predict the label of each test object.
If we want to use the conformal predictor based on the support vector method,
as described in Section \ref{sec:conformal},
we will have to find the set of the Lagrange multipliers
for each test object and for each potential label $Y$ that can be assigned to it.
This would involve solving
$k\left|\mathbf{Y}\right|$ essentially independent optimization problems.
Using the nearest neighbours approach
is typically more computationally efficient,
but even it is much slower than the following procedure,
suggested in \cite{papadopoulos/etal:2002a,papadopoulos/etal:2002b}.

Suppose we have an inductive algorithm which,
given a training set (\ref{eq:training-set}) and a new object $x$
outputs a prediction $\hat y$ for $x$'s label $y$.
Fix some measure $\Delta(y,\hat y)$ of difference between $y$ and $\hat y$.
The procedure is:
\begin{enumerate}
\item
  Divide the original training set (\ref{eq:training-set})
  into two subsets:
  the \emph{proper training set}
  $(x_1,y_1),\ldots,(x_m,y_m)$
  and the \emph{calibration set}
  $(x_{m+1},y_{m+1}),\ldots,(x_l,y_l)$.
\item
  Construct a prediction rule $F$ from the proper training set.
\item
  Compute the nonconformity score
  \begin{equation*}
    \alpha_i:=\Delta(y_i,F(x_i)),
    \quad
    i=m+1,\ldots,l,
  \end{equation*}
  for each example in the calibration set.
\item
  For every test object $x_i$,
  $i=l+1,\ldots,l+k$,
  do the following:
  \begin{enumerate}
  \item
    for every possible label $Y\in\mathbf{Y}$
    compute the nonconformity score $\alpha_i:=\Delta(y_i,F(x_i))$
    and the p-value
    \begin{equation*}
      p_Y
      :=
      \frac
      {
        \#\{j\in\{m+1,\ldots,l,i\} \st \alpha_j\ge\alpha_i\}
      }
      {l-m+1};
    \end{equation*}
  \item
    output the prediction sets
    $
      \Gamma^{\epsilon}
      \left(
        x_1,y_1,\ldots,x_{l},y_{l},x_{i}
      \right)
    $
    given by the right-hand side of (\ref{eq:Gamma}).
  \end{enumerate}
\end{enumerate}
This is a special case of ``inductive conformal predictors'',
as defined in \cite{vovk/etal:2005}, Section 4.1.
In the case of classification,
of course,
we could package the p-values as a simple prediction
complemented with confidence (\ref{eq:conf}) and credibility (\ref{eq:cred}).

Inductive conformal predictors are valid in the sense that
the probability of error
\begin{equation*}
  y_{i}
  \notin
  \Gamma^{\epsilon}
  \left(
    x_1,y_1,
    \ldots
    x_l,y_l,
    x_{i}
  \right)
\end{equation*}
($i=l+1,\ldots,l+k$, $\epsilon\in(0,1)$)
never exceeds $\epsilon$
(cf.\ (\ref{eq:error})).
The on-line version of inductive conformal predictors,
with a stronger notion of validity,
is described in \cite{vovk:2002}
and \cite{vovk/etal:2005} (Section 4.1).

The main advantage of inductive conformal predictors
is their computational efficiency:
the bulk of the computations is performed only once,
and what remains to do for each test example
is to apply the prediction rule found at the inductive step,
to apply $\Delta$ to find the nonconformity score $\alpha$ for this example,
and to find the position of $\alpha$ among the nonconformity scores
of the calibration examples.
The main disadvantage is a possible loss of the prediction efficiency:
for conformal predictors,
we can effectively use the whole training set
as both the proper training set and the calibration set.

\section{Conclusion}
\label{sec:conclusion}

This paper shows how many machine-learning techniques
can be complemented with provably valid measures
of accuracy and reliability.
We explained briefly how this can be done
for support vector machines, nearest neighbours algorithms,
and the ridge regression procedure,
but the principle is general:
virtually any (we are not aware of exceptions) successful prediction technique
designed to work under the randomness assumption
can be used to produce equally successful hedged predictions.
Further examples are given in our recent book \cite{vovk/etal:2005}
(joint with Glenn Shafer),
where we construct conformal predictors and inductive conformal predictors
based on nearest neighbours regression, logistic regression,
bootstrap, decision trees, boosting, and neural networks;
general schemes for constructing conformal predictors
and inductive conformal predictors
are given on pp.~28--29 and on pp.~99--100 of \cite{vovk/etal:2005},
respectively.
Replacing the original simple predictions with hedged predictions
enables us to control the number of errors made
by appropriately choosing the confidence level.

\section*{Acknowledgements}

This work is partially supported by MRC
(grant 
``Pro\-te\-o\-mic analysis of the human serum pro\-te\-ome'')
and the Royal Society
(grant ``Efficient pseudo-random number generators'').

\end{document}